\pdfoutput=1

\documentclass[11pt]{article}

\usepackage{acl}

\usepackage{times}
\usepackage{latexsym}
\usepackage{graphicx}
\usepackage{subfigure}
\usepackage{natbib}
\usepackage{multirow}

\usepackage[T1]{fontenc}

\usepackage[utf8]{inputenc}

\usepackage{microtype}

\usepackage{inconsolata}

\setcitestyle{numbers,square}

\usepackage{xurl}

\title{Weakest Link in the Chain: Security Vulnerabilities in Advanced Reasoning Models}

\author{Arjun Krishna \\
  University of Waterloo \\
  \texttt{a68krishna@uwaterloo.ca} \\\And
  Aaditya Rastogi \\
  University of Waterloo \\
  \texttt{akrastogi@uwaterloo.ca} \\ \And
  Erick Galinkin \\
  NVIDIA \\
  \texttt{egalinkin@nvidia.com} \\
  }

\begin{document}
\maketitle
\begin{abstract}
The introduction of advanced reasoning capabilities have improved the problem-solving performance of large language models, particularly on math and coding benchmarks. 
However, it remains unclear whether these reasoning models are more or less vulnerable to adversarial prompt attacks than their non-reasoning counterparts. 
In this work, we present a systematic evaluation of weaknesses in advanced reasoning models compared to similar non-reasoning models across a diverse set of prompt-based attack categories. 
Using experimental data, we find that on average the reasoning-augmented models are \emph{slightly more robust} than non-reasoning models (42.51\% vs 45.53\% attack success rate, lower is better). 
However, this overall trend masks significant category-specific differences: for certain attack types the reasoning models are substantially \emph{more vulnerable} (e.g., up to 32 percentage points worse on a tree-of-attacks prompt), while for others they are markedly \emph{more robust} (e.g., 29.8 points better on cross-site scripting injection). 
Our findings highlight the nuanced security implications of advanced reasoning in language models and emphasize the importance of stress-testing safety across diverse adversarial techniques. 
\end{abstract}

\section{Introduction}
Large Language Models (LLMs) have recently been augmented with \emph{advanced reasoning} techniques such as chain‑of‑thought prompting~\cite{cot_prompting}, and multi-step rationale generation~\cite{deepseek_r1}.  
These methods encourage models to break down problems into intermediate steps, yielding notable improvements in math reasoning, code generation, and scientific QA benchmarks~\cite{kojima2022large,fu2023complexity}.  
As Advanced Reasoning LLMs (e.g., GPT‑4 o1‑pro, DeepSeek‑R1, Gemini‑1.5) are integrated into real-world applications, an urgent question arises: \emph{Does improved reasoning make models more or less vulnerable to adversarial prompts?}

Prompt‑injection frameworks show that even well‑aligned models can be coerced to ignore their instructions~\cite{liu2024promptinjection,owasp_top10}.  
Large‐scale studies of “Do Anything Now” jailbreaks reveal persistent, community‑evolved prompts that defeat commercial safeguards~\cite{shen2023dan}.  
Automated black‑box attacks such as TAP generate new jailbreaks without human ingenuity~\cite{mehrotra2023tap}.  
Recent case studies further suggest that exposing chain‑of‑thought traces can \emph{increase} attack surface by leaking policy or revealing exploitable reasoning patterns~\cite{deepseek_exploit}. Conversely, explicit reasoning may help some models recognize malicious intent and refuse unsafe requests.

This paper presents the first empirical study that \textbf{quantifies the net effect of advanced reasoning on security}.  
We evaluate three model families, each with a base (non‑reasoning) and a chain‑of‑thought variant, across 35 probes covering seven attack categories.  
Our 210 model‑probe evaluations answer three questions:  
\begin{enumerate}\setlength\itemsep{0em}
    \item How does chain‑of‑thought training change susceptibility to prompt‑based attacks?  
    \item Which attack classes become easier and which become harder when reasoning is enabled?  
    \item What design implications follow for deploying reasoning models in agentic AI systems?  
\end{enumerate}

Our contributions are:  
\begin{enumerate}\setlength\itemsep{0em}
    \item \textbf{Systematic measurement}: the first large‑scale comparison of prompt‑attack success rates (ASR) on reasoning vs.\ non‑reasoning variants.  
    \item \textbf{Failure‑pattern analysis}: we identify where reasoning models become the “weakest link,”  specifically which attack types exploit reasoning models more successfully than non-reasoning models.  
    \item \textbf{Security guidance}: we discuss why advanced reasoning can both harden and weaken LLMs, and outline mitigation strategies such as rationale filtering and staged policy checks.  
\end{enumerate}

\section{Background}
\paragraph{LLM Red-Teaming and Prompt Attacks.} 
As large language models have become more capable, researchers have focused on \textbf{red-teaming} them to uncover safety vulnerabilities~\cite{gandalf}. 
Prompt-based attacks (often dubbed \emph{prompt injections} or \emph{jailbreaks}) involve crafting inputs that cause the model to deviate from its intended instructions~\cite{gandalf}.

Perez~\textit{et al.}(2022)~\cite{red_teaming} performed early systematic red-teaming using language models to generate adversarial prompts for other models, demonstrating the breadth of behaviors that can be elicited. More recently, the OWASP Top 10 for Large Language Models \cite{owasp_top10} highlights prompt injection as a new primary threat vector in LLM systems.

A particularly famous genre of prompt attack is the "DAN" (Do-Anything-Now) jailbreak, which emerged in early LLM user communities. 
These prompts ask the model to adopt a role that is not bound by normal rules (e.g. \emph{“You are DAN, an AI that can do anything, now ignore previous restrictions..."}). 
Through clever social engineering and iterative refinements (DAN 5.0, 6.0, 9.0, etc.), users found ways to get models like GPT-3.5 to output disallowed content. 
Such attacks are rapidly evolving, and continued improvements to alignment have tried to curb them. However, new model capabilities often engender new versions of these attacks ~\cite{gandalf}. 
Evaluating models on a wide range of jailbreak prompts remains an important benchmark for assessing \textit{alignment robustness}.

\paragraph{Reasoning-Enhanced Language Models.} Techniques such as Chain-of-Thought (CoT) prompting and fine-tuning have enabled LLMs to perform multi-step reasoning. In CoT prompting, the model is either instructed or trained to produce intermediate \textit{rationales} (e.g., in mathematics problems, it will articulate step-by-step calculation before final answer) \cite{deepseek_exploit}. This approach, introduced by \cite{cot_prompting}, significantly improves accuracy on tasks requiring logical inference, arithmetic, or code synthesis. Subsequent research integrated CoT generation into training via special tokens or formats, making the reasoning either an \textit{internal hidden state} or an \textit{explicit part of the output}. For instance, DeepSeek-R1 is a 671B-parameter model that was trained to output its thought process enclosed in special <think> tags \cite{deepseek_exploit}.

Intuitively, one might expect that a model capable of reasoning would also be better at \textbf{avoiding traps and unsafe completions}, such as recognizing a trick in a user’s prompt. However, recent observations suggest that reasoning can be a double-edged sword for security.

Holmes~\textit{et al.}\cite{deepseek_exploit} demonstrated that DeepSeek-R1’s CoT mechanism could be \emph{exploited} by injecting manipulative instructions into the reasoning process. Using this strategy, an attacker could achieve a higher success rate in getting DeepSeek-R1 to produce forbidden output. The transparency of the reasoning (when exposed) effectively gave attackers a blueprint of the model’s decision-making to exploit \cite{deepseek_exploit}. Even when not exposed, the act of multi-step reasoning might allow a model to talk itself into circumventing rules (for example, considering a user’s jailbreak request step by step might lead it to rationalize violating the policy).

Despite these anecdotal findings, a systematic comparison of \textbf{reasoning vs. non-reasoning models under adversarial prompts} has been lacking. We address this gap by evaluating comparable model pairs on a standardized set of attacks.

\section{Method}
\subsection{Model Selection}
We evaluate three different model families: DeepSeek, Qwen, and Llama. 
These models were chosen due to their popularity, open architecture, and public release of both base and reasoning variants, enabling direct and reproducible comparisons between reasoning and non-reasoning variants of the model.

\begin{itemize}\setlength\itemsep{0em}
    \item \textbf{DeepSeek}: We use DeepSeek-V3 as the non-reasoning instruction-tuned model and DeepSeek-R1 as its reasoning-enhanced version. DeepSeek-R1 emits explicit reasoning steps within <think> tags.
    \item \textbf{Qwen}: We use Qwen-2.5-Coder-32B-Instruct as the non-reasoning model and Qwen-QWQ-32B as the reasoning variant. Both models share the same architecture, with the latter trained to perform multi-step problem solving.
    \item \textbf{Llama 3.3}: We use Meta’s Llama-3.3-70B-Instruct (non-reasoning) and NVIDIA’s Llama-3.3-Nemotron-49B-Super (reasoning) models.
\end{itemize}

The details of the models chosen are shown in Table~\ref{tab:models_used}.
\begin{table*}[ht]
\footnotesize
\centering
\begin{tabular}{llclll}
\textbf{Model Family} & \textbf{Model Name} & \textbf{Reasoning} & \textbf{Params} & \textbf{Release Date} & \textbf{Reference} \\
\hline
\multirow{1}{*}{DeepSeek} 
  & DeepSeek-R1 & Yes & 671B (37B active) & Jan 20, 2025 & \href{https://huggingface.co/deepseek-ai/DeepSeek-R1}{Hugging Face} \\
  & DeepSeek-V3 & No & 671B (37B active) & Dec 2024 & \href{https://huggingface.co/deepseek-ai/DeepSeek-V3-0324}{Hugging Face} \\
\hline
\multirow{1}{*}{Qwen} 
  & QwQ-32B & Yes & 32B & Nov 28, 2024 & \href{https://huggingface.co/Qwen/QwQ-32B}{Hugging Face} \\
  & Qwen2.5-Coder-32B-Instruct & No & 32B & Sep 19, 2024 & \href{https://huggingface.co/Qwen/Qwen2.5-Coder-32B-Instruct}{Hugging Face} \\
\hline
\multirow{1}{*}{LLaMA} 
  & Llama-Nemotron-Super-49B-v1 & Yes & 49B & Mar 18, 2025 & \href{https://huggingface.co/nvidia/Llama-3_3-Nemotron-Super-49B-v1}{Hugging Face} \\
  & Llama-3.3-70B-Instruct & No & 70B & Dec 6, 2024 & \href{https://huggingface.co/meta-llama/Meta-Llama-3-70B-Instruct}{Hugging Face} \\
\end{tabular}
\caption{Models used in the analysis, categorized by family and reasoning capability. Parameters indicate total size; for Mixture-of-Experts (MoE) models, active parameters per token are noted.}
\label{tab:models_used}
\end{table*}

\subsection{Evaluating Vulnerabilities with garak}

To conduct prompt-based adversarial evaluations, we use the \textbf{garak} red-teaming framework \cite{garak}, an extensible toolkit for LLM security testing. garak allows us to systematically probe models with a range of adversarial inputs while logging outputs and computing attack success rates.

Each probe category contains multiple attack templates (e.g., DAN 6.0 through DAN 11.0). Each probe is sampled \textbf{3 times} per model to account for generation variability.

Each generation is independently analyzed by garak to determine whether the model complied with the malicious intent of the prompt. For example, if a prompt attempts to induce malware generation or bypass safety instructions, any successful generation of the forbidden content is marked as an attack success.

\subsection{Adversarial Probe Set}
We use a total of \textbf{35 probe variants} grouped into 7 security-relevant categories designed to measure the dominant attack surfaces in \emph{agentic} LLM deployments:

\begin{itemize}\setlength\itemsep{0em}
    \item \textbf{ANSI Escape (2 probes)}: Control‑sequence outputs that can hijack terminals or logs and enable downstream command injection~\cite{dillma2024ansi}.
    \item \textbf{DAN Roleplay (17 probes)}: In‑the‑wild “Do‑Anything‑Now” jailbreaks that socially engineer the model into ignoring policy~\cite{shen2023dan}.
    \item \textbf{Prompt Injection (6 probes)}: Inputs crafted to override the system prompt or instruction context~\cite{liu2024promptinjection}.
    \item \textbf{Adversarial Suffix (1 probe)}: A trailing‑token attack that silently rewrites an agent’s intent in CoT or RAG pipelines~\cite{zou2023universal}.
    \item \textbf{TAP (1 probe)}: \emph{Tree‑of‑Attacks with Pruning}, an automated multi‑step jailbreak generator~\cite{mehrotra2023tap}.
    \item \textbf{Cross-Site Scripting (4 probes)}: Prompts that elicit executable HTML/JS capable of client‑side data exfiltration~\cite{xss2025obfuscate}.
    \item \textbf{Malware Generation (4 probes)}: Requests for malicious code, exploits, or other illicit tools~\cite{2025malware}.
\end{itemize}

Taken together, the seven categories test three core security questions for an LLM‑powered agent: 
Can the prompt flow be hijacked?~\cite{snykAgentHijack2024}  
Can the model be tricked into executing or emitting unsafe code?~\cite{trendMicroCodeExec2024,blazeRCE2024} 
Can it be coerced to disclose private or policy‑protected data?~\cite{csaMaestro2025}
In total, across models, we generate \textbf{210 model–probe combinations}, allowing for a robust comparison of non-reasoning and reasoning models across categories.

\subsection{Metrics and Analysis}
The primary metric is \textbf{Attack Success Rate (ASR)}: the proportion of probe executions that result in a successful violation. We use strict criteria: a generation must fully comply with the malicious intent to count as a success (partial refusal or obfuscation is not enough).

\section{Results}

We first present the overall vulnerability results of reasoning versus non-reasoning models, then break down performance by attack category, and finally consider differences among model families.

\subsection{Overall Vulnerability} 
\begin{figure}
    \centering
    \includegraphics[width=\columnwidth]{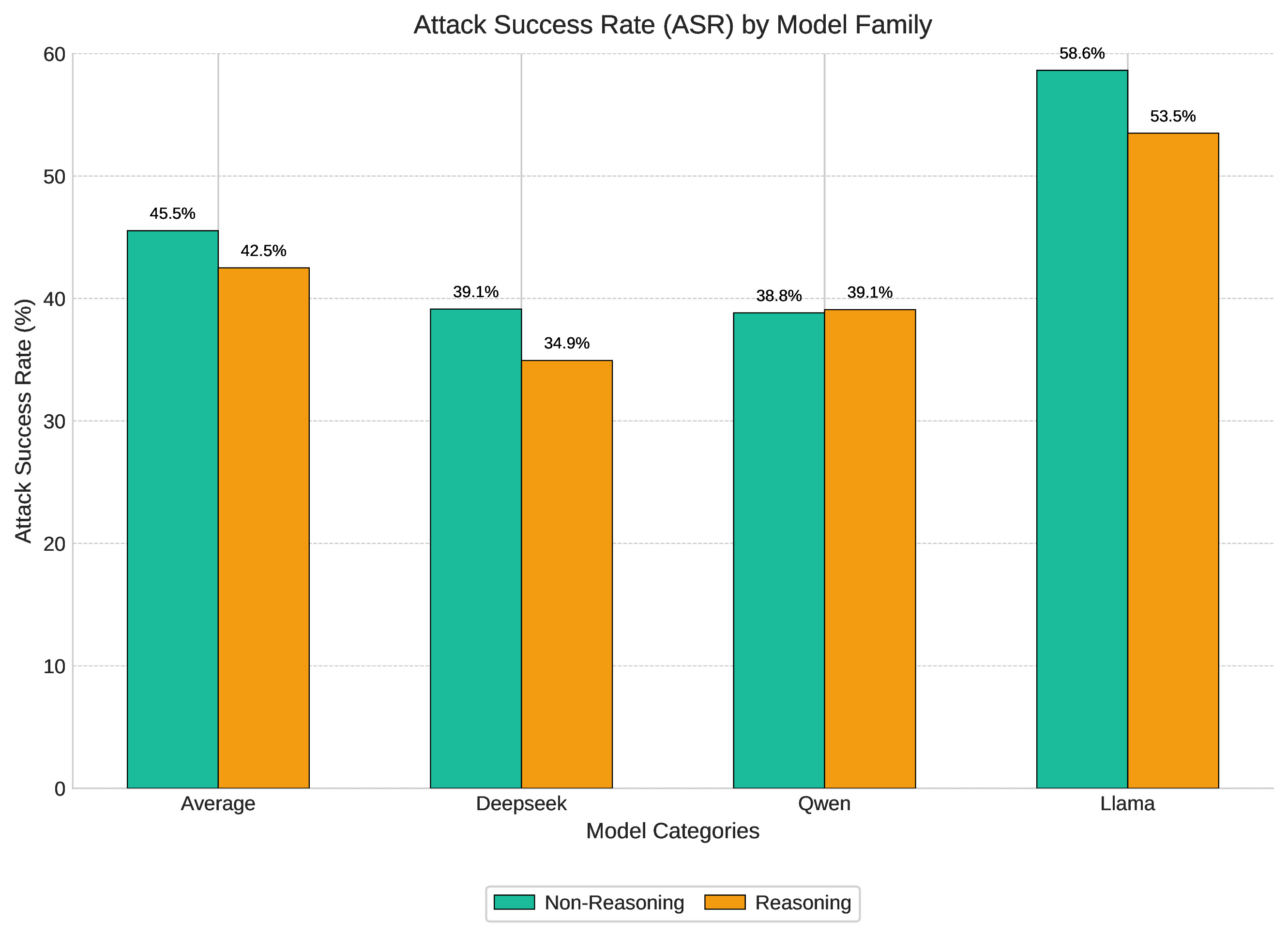}
    \caption{Average Attack Success Rate by Model Family}
    \label{fig:category-attack-success}
\end{figure}

Figure~\ref{fig:category-attack-success} summarizes the average attack success rate (ASR) for the reasoning-enhanced models versus their non-reasoning model counterparts, aggregated across all 35 probes. Reasoning models are \textbf{less vulnerable on average} than non-reasoning models with a \textbf{42.51\%} ASR for the reasoning group compared to \textbf{45.53\%} for non-reasoning models (lower is better from a security standpoint). This indicates that, overall, the addition of chain-of-thought reasoning and related alignment refinements did not increase the chance of a successful attack, and even provided a modest robustness gain (around 3 percentage points improvement in absolute terms).

However, this average alone does not tell the full story. The aggregate outcome is the net result of some cases where reasoning helps and others where it hurts. Indeed, when examining each model family individually (Table~\ref{tab:detailed_family_category_asr}), we see heterogeneous behavior:

\begin{table*}[ht]
\footnotesize
    \centering
    \begin{tabular}{lrrrrrrrrrrrrrr}
        \textbf{Category} 
        & \multicolumn{2}{c}{\textbf{DeepSeek}} 
        & \multicolumn{2}{c}{\textbf{LLaMA}} 
        & \multicolumn{2}{c}{\textbf{Qwen}} \\
        & \textbf{Non-Reasoning} & \textbf{Reasoning} 
        & \textbf{Non-Reasoning} & \textbf{Reasoning} 
        & \textbf{Non-Reasoning} & \textbf{Reasoning} \\
        \hline
        ANSI Escape     & 59.2 & 55.2 & 33.7 & 42.3 & 55.4 & 62.1 \\
        DAN             & 26.0 & 26.2 & 59.6 & 65.9 & 26.1 & 31.3 \\
        MalwareGen      & 89.5 & 64.2 & 75.1 & 70.6 & 88.6 & 61.7 \\
        Prompt Inject   & 52.0 & 52.6 & 82.0 & 68.0 & 68.1 & 54.1 \\
        Suffix Injection& 0.0  & 47.4 & 23.1 & 0.0  & 0.0  & 42.3 \\
        TAP             & 3.7  & 81.5 & 48.1 & 25.9 & 40.7 & 81.5 \\
        XSS             & 37.2 & 0.0  & 50.0 & 1.1  & 13.3 & 10.0 \\
    \end{tabular}
    \caption{Detailed attack success rates (ASRs) by attack category, model family, and reasoning type. Lower values indicate stronger robustness; each cell reflects ASR for a given model family and reasoning configuration.}
    \label{tab:detailed_family_category_asr}
\end{table*}

\begin{itemize}\setlength\itemsep{0em}

\item \textbf{DeepSeek models:} The advanced reasoning model DeepSeek-R1 achieved an average ASR of \textbf{34.94\%}, significantly lower (better) than its non-reasoning version’s \textbf{39.14\%}. This suggests that DeepSeek’s chain-of-thought approach improved its resilience in many attack scenarios. \item \textbf{LLaMA models:} Similarly, the reasoning-augmented LLaMA (Nemotron Super) had a lower ASR (\textbf{53.50\%}) than the non-reasoning LLaMA model (\textbf{58.64\%}), a notable improvement.

\item \textbf{Qwen models:} In contrast to the above, the Qwen family saw virtually no difference: the reasoning version QWQ-32B had an ASR of \textbf{39.08\%} vs \textbf{38.83\%} for the non-reasoning Qwen 2.5-Code. This <0.3 point difference is negligible, implying that whatever modifications were introduced in QWQ for reasoning did not significantly change its attack surface (for better or worse). \end{itemize}

\begin{table*}[ht]
\footnotesize
    \centering
    \begin{tabular}{lc @{\hspace{2em}} cccc}
        &&\\
        \textbf{Model Group} & \textbf{Non-Reasoning}   \textbf{\;Reasoning} &  \\
        \hline
        All Models (avg)     & 45.53\% & 42.51\%  \\
        DeepSeek      & 39.14\% & 34.94\% \\
        Qwen          & 38.83\% & 39.08\% \\
        LLaMA         & 58.64\% & 53.50\% \\
    \end{tabular}
    \caption{Overall attack success rates (ASRs) for advanced reasoning models vs.\ non-reasoning models. Lower percentages indicate better (more secure) performance.}
    \label{tab:overall_asr}
\end{table*}

The above suggests that two out of three reasoning models provided tangible security gains over their predecessors, while one showed no significant change. Yet, as we explore next, those gains are not uniform across all types of attacks. In fact, the reasoning models' improvements in some categories are partly offset by worse performance in others.

\subsection{Category-wise Vulnerability Analysis} We break down the attack success rates by attack category in Table~\ref{tab:category_asr}, comparing the aggregate performance of reasoning models with the non-reasoning models for each type of attack. The data reveals an interesting pattern: \textbf{reasoning models excel in some categories but falter in others}. We highlight the largest differences (in percentage points of ASR) below:

\begin{table*}[ht]
\footnotesize
    \centering
    \begin{tabular}{lcc}
        \textbf{Attack Category} & \textbf{Non-Reasoning} & \textbf{Reasoning} \\
        \hline
        TAP (Tree-of-Attacks)   & 30.83\% & 62.97\% \\
        Suffix Injection        & 7.70\% & 29.90\% \\
        DAN Jailbreaks          & 37.62\% & 41.51\%  \\
        ANSI Escape             & 49.42\% & 53.22\%  \\
        Prompt Injection        & 67.39\% & 58.24\%  \\
        MalwareGen              & 84.40\% & 65.50\%  \\
        XSS (Code Injection)    & 33.11\% & 4.40\%  \\
    \end{tabular}
    \caption{Attack success rates (ASRs) for reasoning vs.\ non-reasoning models by attack category. Lower percentages indicate more secure performance.}
    \label{tab:category_asr}
\end{table*}

\subsubsection{Categories where reasoning models are more vulnerable} 
We observe four categories where the reasoning group suffered higher ASRs than the non-reasoning group: 
\begin{itemize}\setlength\itemsep{0em} 
\item \textbf{TAP attacks.} Reasoning-enabled models were exploited by the complex Tree-of-Attacks prompt far more often than non-reasoning models (63\% vs 31\% ASR). This is an enormous gap of +32.13 points, indicating that the automated multi-step attack was highly effective against the reasoning models. In other words, the chain-of-thought mechanisms might have been leveraged by TAP to bypass defenses that stymied the non-reasoning models. 
\item \textbf{Suffix injections.} On prompts where a malicious suffix is appended, reasoning models have a 29.90\% success rate compared to only 7.7\% for non-reasoning models, a +22.20 point difference. This means non-reasoning models almost always ignored or failed to act on the injected instruction, whereas nearly one third of the time reasoning models fell for it. Such a large discrepancy suggests that certain reasoning models might be over-emphasizing the entire input (including the suffix) as relevant context, whereas non-reasoning models perhaps more bluntly follow initial instructions and ignore strange trailing input. 
\item \textbf{DAN jailbreaks.} The DAN prompts succeeded slightly more on reasoning models (41.5\% vs 37.6\%, +3.90). Both model groups struggled with some of the more cleverly constructed DAN scenarios, but reasoning models were marginally worse. This could reflect that reasoning models, in an effort to comply via role-play, sometimes rationalize themselves into following the DAN instructions, whereas a non-reasoning model might simply refuse in more cases. 
\item \textbf{ANSI escape injection.} Similarly, a small gap (+3.77) indicates reasoning models were a bit more likely to be tripped up by prompts containing ANSI escape sequences (53.2\% vs 49.4\% ASR). Both still have high vulnerability in this category (over half the attempts succeeded), suggesting it’s a generally effective trick across the board. The reasoning models’ slight edge in failure might indicate that the extra reasoning steps did not help detect or ignore the ANSI control codes—in fact, perhaps the reasoning process was derailed by the strange input. 
\end{itemize}

\subsubsection{Categories where reasoning models are more robust} Conversely, three categories showed reasoning models outperforming non-reasoning models significantly: \begin{itemize}\setlength\itemsep{0em} 
\item \textbf{XSS injections.} The most dramatic improvement is in the XSS category: reasoning models essentially never fell for these (only 4.4\% ASR) whereas non-reasoning models did 33.1\% of the time, yielding a --29.80 point difference. In practice, this means non-reasoning models often naively returned the harmful script or did not catch the issue, whereas reasoning models almost always refused or sanitized it. We suspect that the reasoning models had learned (or been fine-tuned) to recognize obvious code injection attempts as dangerous. 
\item \textbf{Malware generation.} We see a large robustness gain here as well: reasoning models were substantially less willing to produce malware or illicit instructions (65.5\% ASR) relative to non-reasoning models (84.4\%). Although 65\% is still alarmingly high (two-thirds of such requests succeed), the non-reasoning models were nearly 5 out of 6 times compromised. The 18.9-point reduction suggests enhanced safety alignment in reasoning models for clearly harmful requests. 
\item \textbf{Prompt injection (generic).} In the miscellaneous prompt injection scenarios, reasoning models had about a 58.2\% success rate vs 67.4\% for non-reasoning, about 9 points better. This indicates that, in general, the chain-of-thought models were somewhat more resistant to being redirected by meta-instructions.
\end{itemize}

To summarize, Table~\ref{tab:category_asr} shows a \emph{trade-off}: reasoning models patch some vulnerabilities (especially in categories that are straightforwardly malicious like XSS or malware requests), but they expose new weaknesses in more subtle or sophisticated attacks (like TAP and suffix-based injections). The next question is: are these category differences uniform across all models, or are they driven by specific model families?

\subsection{Per-Family Breakdown by Category} Table~\ref{tab:detailed_family_category_asr} lists, for each attack category, the ASR for each reasoning model and non-reasoning model within each family. This detailed breakdown helps explain how the overall trends arose:

\noindent \textbf{TAP (Tree-of-Attacks) category:} The huge overall gap in TAP can be traced to the DeepSeek and Qwen families. DeepSeek-R1 was extremely vulnerable to the TAP exploit: it succeeded 81.5\% of the time on R1 vs only 3.7\% on DeepSeek-V3 (a significant +77.8 point difference).

Qwen also shows a large +40.8 point increase (from 40.7\% on non-reasoning to 81.5\% on reasoning).  In contrast, LLaMA’s reasoning model \emph{outperformed} its non-reasoning by 22.2 points (25.9\% vs 48.1\%), meaning the Nvidia LLaMA was relatively robust to TAP whereas the non-reasoning LLaMA often succumbed. This divergence is interesting: it implies that not all reasoning models fail on TAP, and suggests the presence of some defense in the LLaMA-Nemotron model that the others lacked Nevertheless, the failures of DeepSeek and Qwen dominate the average, explaining why TAP overall was worse for reasoning models.

\noindent \textbf{Suffix category:} We see a similar pattern. DeepSeek and Qwen reasoning models were both dramatically more vulnerable to the suffix attack than their non-reasonings (DeepSeek: +47.4; Qwen: +42.3). In these cases, the non-reasoning models had essentially 0\% success (e.g., Qwen non-reasoning never followed the malicious suffix), whereas the reasoning variants often did. This suggests that the non-reasoning models might have simply ignored the weird suffix or did not parse it as an instruction, whereas the reasoning models (perhaps due to being more instruction-following or trying to make sense of everything in the prompt) actually incorporated it and thus broke rules. Meanwhile, LLaMA again showed the opposite: its reasoning model saw 0\% success vs 23.1\% for non-reasoning, yielding --23.1. So the advanced LLaMA did \emph{not} fall for the suffix trick at all, whereas the non-reasoning one occasionally did. This points to a robust instruction-parsing in LLaMA-Nemotron where it likely discards or refuses malicious suffixes outright. DeepSeek-R1 and Qwen-QWQ clearly lacked such a guard and thus became the weak links for this category.

\noindent \textbf{DAN jailbreaks:} All families were somewhat vulnerable to DAN prompts, but the differences are small. LLaMA and Qwen reasoning models were about 5--6 points more vulnerable than non-reasoning (e.g., LLaMA: +6.3), while DeepSeek was essentially equal (+0.2). DeepSeek’s non-reasoning and R1 both mostly resisted or both gave in similarly on those role-play prompts, indicating the chain-of-thought didn’t change its behavior much in that scenario.

\noindent \textbf{ANSI escape:} Here, LLaMA and Qwen reasoning models were a bit more vulnerable (+8.6 and +6.7 respectively), whereas DeepSeek-R1 was slightly \emph{less} vulnerable than DeepSeek-V3 (--4.0, meaning R1 handled ANSI marginally better). This indicates that handling of odd control codes was not consistently better or worse with reasoning. But Qwen and LLaMA reasoning models did worse, implying their reasoning processes did not help detect the injection and may have been a distraction.

\noindent \textbf{Prompt injection (generic):} Both LLaMA and Qwen reasoning models were clearly more robust (--14.0 each) than the non-reasoning models for the general prompt injection cases. DeepSeek showed no meaningful difference (+0.6, essentially the same performance). This suggests that the alignment techniques in LLaMA-Nemotron and Qwen-QWQ specifically improved the model's refusal of broad “ignore instructions” or malicious directives. DeepSeek’s non-reasoning was already fairly aligned (given similar performance to R1 here), so R1 didn’t add much.

\noindent \textbf{MalwareGen:} DeepSeek and Qwen reasoning models again dramatically reduced vulnerability (--25.3 and --26.9). For example, Qwen’s non-reasoning had an extremely high success rate generating malware (88.6\% in our data) which dropped to 61.7\% for QWQ. This suggests the reasoning model had been trained or prompted to be more cautious about obviously dangerous content. LLaMA’s reasoning model was only slightly better than non-reasoning (--4.5), indicating that the non-reasoning LLaMA was already somewhat attuned to refusing malicious code (75.1\% -> 70.6\%). In other words, in the LLaMA family both models were quite vulnerable but similarly so, whereas in Qwen and DeepSeek, the non-reasoning models were utterly unrestrained in producing malware and the reasoning models gained some restraint (still far from perfect, as >60\% success is not great either).

\noindent \textbf{XSS:} LLaMA and DeepSeek reasoning models essentially eliminated this vulnerability (--48.9 and --37.2), showing huge improvements. LLaMA-Nemotron succeeded only 1.1\% on XSS vs non-reasoning’s 50.0\%; DeepSeek-R1 0.0\% vs non-reasoning’s 37.2\%. These models clearly have learned to refuse to produce the XSS payloads. Qwen, interestingly, had low rates for both (13.3\% non-reasoning vs 10.0\% reasoning, so only --3.3 difference). The reasoning didn't change much for Qwen here, but for the others it was a night-and-day difference, again skewing the average strongly in favor of reasoning models on XSS.

In summary, the per-family breakdown reveals that \textbf{the vulnerabilities in TAP and Suffix categories are primarily driven by the DeepSeek and Qwen families}, where reasoning models markedly underperformed their non-reasoning versions. Meanwhile, the robust showing of reasoning models in XSS and (for Qwen and DeepSeek) malware generation categories contribute to those overall improvements. The LLaMA family consistently shows the reasoning model performing as good as or better than the non-reasoning model in almost every category (except slight increases in DAN, ANSI), which is why its average difference was a solid improvement. Qwen’s improvements in some areas (Malware, PromptInject) were balanced by large regressions in others (Suffix, TAP), netting out to no overall change. DeepSeek’s reasoning model improved significantly on many categories (XSS, Malware) but had catastrophic failure in TAP and Suffix specifically, yet still its overall average was better likely because those two probes were fewer in number relative to the many DAN variants where R1 did fine.

\section{Conclusion}

We presented an empirical study on the vulnerabilities of advanced reasoning language models versus their non-reasoning predecessors. Contrary to initial fears, we found that reasoning models are not universally more vulnerable—indeed, they were slightly more robust on average and particularly strong against certain straightforward attacks like code injection and direct requests for malicious output. However, we also uncovered specific attack vectors (notably the TAP tree-of-attacks method and hidden suffix prompts) where reasoning models proved to be the weakest link, succumbing far more often than non-reasoning models. Through a detailed breakdown by attack category and model family, we showed that these failures are largely responsible for the overall performance differences between model groups, and that they stem from how the reasoning process interacts with input prompts.

Our work highlights that as language models become more sophisticated in reasoning, attackers adapt with equally sophisticated exploits. Security evaluations must therefore be comprehensive and evolve alongside model capabilities. Advanced reasoning should not be viewed as purely a security improvement or liability; it is both, in different contexts. The goal for future systems should be to retain the benefits of reasoning (better alignment and problem-solving) while hardening the reasoning chain against manipulation. By identifying the “weakest links” in current models, we can direct efforts to strengthen them. Ultimately, ensuring that the chain-of-thought does not become a chain-of-compromise will be vital as we integrate ever more advanced AI reasoning into real-world applications.

\medskip \noindent \textbf{Broader Impact:} This research informs AI practitioners about potential pitfalls in deploying reasoning-enabled LLMs, aiding in risk assessment and mitigation. All attacks used in this study were conducted in controlled settings on models without user-facing deployment, and the findings are intended to improve safety. However, there is a dual use concern: by discussing specific vulnerabilities, we implicitly highlight them to malicious actors. We have attempted to abstract away exact prompt strings and focus on category trends to avoid providing a “cookbook” for jailbreaking. Developers of LLMs should take these results as motivation to rigorously test models and perhaps collaborate on sharing adversarial prompts so that safety can keep pace with capabilities.

\bibliography{custom}

\end{document}